\documentclass[10pt, conference, compsocconf]{IEEEtran}
\pdfoutput=1
\usepackage{amsmath}
\usepackage{amssymb}
\usepackage{latexsym}
\usepackage{graphicx}
\usepackage{subfigure} 
\usepackage{url}
\usepackage{todonotes}
\usepackage[hyperfootnotes=false,hidelinks=true]{hyperref}
\title{Monitoring Term Drift Based on Semantic Consistency in an Evolving Vector Field}

\author{\IEEEauthorblockN{Peter Wittek\IEEEauthorrefmark{1}\IEEEauthorrefmark{2},
S\'andor Dar\'anyi\IEEEauthorrefmark{2},
Efstratios Kontopoulos\IEEEauthorrefmark{3}, 
Theodoros Moysiadis\IEEEauthorrefmark{4} and
Ioannis Kompatsiaris\IEEEauthorrefmark{3}}
\IEEEauthorblockA{\IEEEauthorrefmark{1}ICFO-The Institute of Photonic Sciences,
08860 Castelldefels, Barcelona, Spain}
\IEEEauthorblockA{\IEEEauthorrefmark{2}University of Bor{\aa}s,
50190 Bor{\aa}s, Sweden}
\IEEEauthorblockA{\IEEEauthorrefmark{3}Information Technologies Institute,
Centre for Research \& Technology - Hellas, 57001 Thessaloniki, Greece}
\IEEEauthorblockA{\IEEEauthorrefmark{4}Institute of Applied Biosciences,
Centre for Research \& Technology - Hellas, 57001 Thessaloniki, Greece}}

\begin{document}
\maketitle
\begin{abstract}
Based on the Aristotelian concept of potentiality vs. actuality allowing for the study of energy and dynamics in language, we propose a field approach to lexical analysis. Falling back on the distributional hypothesis to statistically model word meaning, we used evolving fields as a metaphor to express time-dependent changes in a vector space model by a combination of random indexing and evolving self-organizing maps (ESOM). To monitor semantic drifts within the observation period, an experiment was carried out on the term space of a collection of 12.8 million Amazon book reviews. For evaluation, the semantic consistency of ESOM term clusters was compared with their respective neighbourhoods in WordNet, and contrasted with distances among term vectors by random indexing. We found that at 0.05 level of significance, the terms in the clusters showed a high level of semantic consistency. Tracking the drift of distributional patterns in the term space across time periods, we found that consistency decreased, but not at a statistically significant level. Our method is highly scalable, with interpretations in philosophy.
\end{abstract}

%%%%%%%%%%%%%%%%%%%%%%%%%%%%%%%%%%%%%%%%%%%%%%%%%%%%%%%%%%%%%%%%%%%%%%%
\section{Introduction}
The modeling of semantic content for statistical analysis, prominently by means of computational and theoretical linguistics, has been quietly inspired by physics and chemistry over the past two decades. Strictly on a metaphoric basis, the idea was to compare language as a rule-based system to domains of natural science as like systems for innovative model design. Such endeavours typically go back to two kinds of physical phenomena, i.e.\ attraction acting on its own like in gravity, a force with non-polar roots, vs. a system of attraction and repulsion based in polarity like electromagnetism. Both word meaning and sentence meaning show statistical behaviour compliant with the idea of non-polar~\cite{daranyi12connecting,cooper1999linguistic} vs. polar~\cite{beeferman1997mod,yuret1998discovery,renouf2007search} binding forces, allowing for  latent analytic thinking for category building in many areas, including natural language processing~\cite{taira2007field}, bioinformatics networks~\cite{liao2009isorankn,yang2009citationrank}, quantum theory~\cite{blacoe2013quantum}, or chemometry~\cite{bylesjo2006opls}. Clearly, similarity of meaning as an attractor vs. difference of meaning as a repellent are organizing principles of conceptual processing one cannot ignore, and an interesting way ahead is to explore the implications of this observation.

Below we will consider word semantics as the ``behaviour'' of linguistic signs of a dual nature, i.e.\ intertwined form and content, leading to the emergence of conceptual categories over objects, and ultimately to the applicability of artificial neural networks~\cite{kohonen2001som} for machine learning. Mathematical ``energy''~\cite{bruce2001biomedical,wang2001indexing} and machine learning are related, the latter often being based on minimizing a constrained multivariate function such as a loss function. Concepts in feature space ``sit'' at energy minima, representing the cost of a classification decision as an energy minimizing process. This suggests that machine learning must identify concepts with such minima, and since potential energy in physics is carried by a field or a respective topological mapping, concepts naturally have something to do with energy as work capacity. As this general process is practically isomorphic with the theory of reaction paths over a potential hypersurface leading to the probabilistic composition of chemical compounds in computational chemistry~\cite{mezey1987potential,levine2009quantum}, we believe that evolving fields as a metaphor to simulate category formation by semantic content is a legitimate approach. Furthermore, attractor networks~\cite{amit1992modeling, cooper1999linguistic} establishing a quasi-continuous field~\cite{widdows2014reasoning} and capable of processing both word and sentence meaning link the above considerations with the study of neural networks.

Key to our current line of thought is the \emph{semantic continuity hypothesis}, i.e.\ the assumption that any vocabulary modelled by term space consists of both actual and potential word content, the former mapped to observable locations, the latter filling in the so-called ``lexical gaps'' between them. Linguistics offers innumerable examples for the existence of such gaps where a language lacks spelled out, i.e.\ actualized content in contrast to another one. This continuity is best modelled as an evolving field, with both actual and potential word content constantly dislocated over time. Due to such dislocations, both the actual positions and their embedding potential contexts may change, offering a rich texture of semantic substance quasi charging actual term locations vs. discharged potential ones. The same line of thought applies to the vector space model of sentence content~\cite{cohen2012discovery}. Given timestamped data, one can measure such dislocations, called the \emph{semantic drift}, an important indicator of ongoing language change~\cite{baker08languageChange,deo2015diachronic}, prominently affecting the monitoring of novelties in document indexing terminology.

We aim to demonstrate the following objectives:
\begin{enumerate}
\item We are interested in evaluating semantic consistency within single time periods of an evolving data set.
\item We would like to see if semantic drift can be detected by analysing the change in semantic consistency.
\end{enumerate}

%%%%%%%%%%%%%%%%%%%%%%%%%%%%%%%%%%%%%%%%%%%%%%%%%%%%%%%%%%%%%%%%%%%%%%%%%%%%%%
\section{Background}

In what follows we introduce four considerations leading to our methodology underlying the experiment design.

%%%%%%%%%%%%%%%%%%%%%%%%%%%%%%%%%%%%%%%%%%%%%%%%%%%%%%%%%%%%%%%%%%%%%%%%%%%%%%
\subsection{Semantic similarity}

As object or feature categorization by neural networks depends on the concept of similarity as a fundamental ``binding force'', we briefly review \emph{measures of semantic relatedness} (MSR) to express thematic coherence~\cite{turney2010frequency}. In linguistics relevant for text processing, there are two prominent theories of word meaning, the distributional hypothesis~\cite{harris70:distr}, and the referential theory of word semantics~\cite{frege1892sb}. According to the first, meaning depends on word use, i.e.\ is contextual, whereas for the second one, it is referential, i.e.\ goes back to convention expressed e.g. by definitions in ontology entries. Because habitual word use as context clearly implies agreements about the sense in which certain word forms are being used in certain contexts, there is a dependency between the two approaches.

Automated systems assign a score of semantic relatedness to a given pair of terms calculated from a relatedness measure. The absolute score itself is typically irrelevant on its own; what is important is that the measure assigns a higher score to term pairs which humans think are more related and comparatively lower scores to term pairs that are less related~\cite{mohammad2005dmp}.

Distributional similarity and its predecessors go back a long way, building on the notion of term dependency and structures derived therefrom~\cite{morris2003tra}. The underlying distributional hypothesis is often cited for explaining how word meaning enters information processing~\cite{karlgren2001wu}. Before attempts to utilize lexical resources for the same purpose, this used to be the sole source of word semantics in information retrieval, inherent in the exploitation of term occurrences -- most notably, in the term frequency-inverse document frequency (TFIDF) measure -- and co-occurrences~\cite{gallant91,peat1991ltc,schutze97}, including multiple-level term co-occurrences~\cite{kontostathis2006ful}. On the other hand, the referential approach relies these days on lexical resources. A lexical resource in computer science is a structure that captures semantic relations among terms, quasi ``charging'' word occurrences in context with external information. 

The reason for combining the two approaches is that statistical techniques typically suffer from the sparse data problem: they perform poorly when the terms are relatively rare. Hybrid methods attempt to address this problem by supplementing sparse data with information from a lexical database~\cite{resnik1995uic,jiang1997ssb}. In a semantic network, to differentiate between the weights of edges connecting a node and all its child nodes, one needs to consider the link strength of each specific child link. This is a situation in which corpus statistics can contribute. The following types of resources are commonly used in measuring semantic similarity between terms: dictionaries~\cite{lesk1986}, semantic networks, such as WordNet~\cite{miller1995wordnet}, thesauri modelled on Roget's Thesaurus~\cite{morris1991lcc}, and ontologies.

%%%%%%%%%%%%%%%%%%%%%%%%%%%%%%%%%%%%%%%%%%%%%%%%%%%%%%%%%%%%%%%%%%%%%%%%%%%%%%
\subsection{Semantic fields}
We find the tradition of using a combination of two planes to describe a phenomenon in several disciplines. E.g. the general practice of evaluating the effectiveness of information retrieval and text categorization models by measures like recall, precision, accuracy, and many more~\cite{turney2010frequency}. There is ongoing work to build semantic spaces from distributional vs. compositional semantics~\cite{pado2007dependency,erk2008structured}, representing both word and sentence meaning as locations in high-dimensional space where for phrase or sentence component binding, recursive matrix-vector spaces~\cite{socher2012semantic}, the tensor product~\cite{baroni2010distributional,blacoe2013quantum,grefenstette2013multi}, or circular holographic reduced representation are routinely used~\cite{cohen2012discovery}. In these models, the representation of semantic content in documents is compared to an ideal state of language use, provided by the human standards of interpretation inherent in the evaluation method~\cite{erk2013measuring}. Using geometry or probability as a vehicle of meaning, i.e.\ building a new medium of language, aims at maximizing similarity between the human standard and its statistical reconstruction. This hypothetic original, a correlate of spoken language called a mental state or internal state in neuroscience~\cite{elman2004alternative}, recalls the ``language of thought hypothesis'' in philosophy~\cite{fodor1975language}, also called \emph{mentalese}. A joint element in the above is that whereas language as a mental phenomenon is assumed to be continuous, its uttered or mathematically modelled representations are discrete.

The same duplicity returns as ``hidden metaphysics'' in traditional mentalist and more recent generalist-universalist theories about language: language is but a tool operated by something deeper -- thought, reason, logic, cognition -- which functions in line with biological-neurological mechanisms common to all human beings~\cite{house2000linguistic}. Moreover, a linguistic school of thought orthogonal to the above theories, called Neo-Humboldtian field theories of word meaning, goes back to the same dual model where discrete distributions of related content called lexical or semantic fields, based on language use, are underpinned by the assumption of conceptual fields in the mind. Then, the lexical field of related words is only an outward manifestation of the underlying conceptual field so that the sum total of conceptual fields describes one's world view~\cite{trier1934sprachliche}. In yet another unrelated school of thought, Saussure's structural linguistics, language (\emph{langue}) is a mental grammar with a rule set specifying ideal content pronunciation, whereas speech (\emph{parole}) stands for the exemplification of those rules~\cite{desaussure1916course}.

An important symptom of lexical fields is that regions of related content are separated by \emph{lexical gaps}. These are nonexistent names for things where one could exist by rules of a particular language, and indicate possible conceptual distinctions not mapped to actual language use, such as mother's father (Swedish \emph{morfar}) vs. father's father (Swedish \emph{farfar}), both called \emph{grandfather} in English, or father's brother (Swedish \emph{farbror}) not distinguished from mother's brother (Swedish \emph{morbror}), both called \emph{uncle}. Such language-specific discontinuities of semantic content play a prominent role in our methodology.

The assumption that products of the mind are continuous while their mapping to spoken language is discrete goes back ultimately to Aristotle's \emph{Metaphysics}. In this, existence or reality is described as the sum total of two components, conceivable potentiality (\emph{dynamis}) plus observable-measurable actuality (\emph{energeia}). These are names for the latent vs. manifest capacity of existents to induce change. Therefore in our current thinking, existence consists of two layers, potentiality (a continuum) and actuality (a discrete distribution sampling the former). Importantly, one ascribes a field nature to mental experience because of the potentiality layer which we indirectly perceive by the actualized values of events.

%%%%%%%%%%%%%%%%%%%%%%%%%%%%%%%%%%%%%%%%%%%%%%%%%%%%%%%%%%%%%%%%%%%%%%%%%%%%%%%
\subsection{Measuring semantic consistency}

For any model departing from the idea of similarity between instances in a semantic field, a logical next question is, how coherent are the groups of instances in that field? Relating term similarity and semantic consistency, the \emph{domain restriction hypothesis} answers that question~\cite{gliozzo2007domain}. Based on the filtering away of extracted but false sense relations, semantically related terms extracted from a corpus tend to be semantically coherent. To this end, \emph{semantic domains} are used as filters by integrating pattern-based and distributional approaches to capture two characteristic properties of semantic relations: 

\begin{itemize}
\item Syntagmatic properties: if two terms X and Y are in a given relation, they tend to co-occur in texts, and are mostly connected by specific lexical-syntactic patterns (e.g., the pattern ``X is a Y'' connects terms in \emph{is-a} relations). This aspect is captured using a pattern-based approach;
\item Domain properties: if a semantic relation among two terms X and Y holds, both X and Y should belong to the same semantic domain (i.e.\ they are semantically coherent), where semantic domains are sets of terms characterized by very similar distributional properties in a (possibly domain specific)
corpus.
\end{itemize}

This approach is detailed in~\cite{gliozzo2009semantic}. On the other hand, in a recent reincarnation, semantic consistency is a new distant supervision method which can identify reliable instances from noisy instances by inspecting whether an instance is located in a semantically consistent region. One way to find out is to first model the local subspace around an instance as a sparse linear combination of training instances, then estimate the semantic consistency by exploiting the characteristics of the local subspace~\cite{hansemantic}.

%%%%%%%%%%%%%%%%%%%%%%%%%%%%%%%%%%%%%%%%%%%%%%%%%%%%%%%%%%%%%%%%%%%%%%%%%%%%%%
\subsection{Semantic drifts}

In the context of Semantic Web dynamics~\cite{antoniou2011semantic}, there is a growing body of literature about the semantic drift~\cite{lauriston1995criteria,gulla2011concept}, the language-related version of abrupt parameter value changes in data mining called concept drifts~\cite{delany2005case,wang2011concept,ross2012exponentially,goncalves2013rcd}. By semantic drift we mean how the features of ontology concepts gradually change as their knowledge domain evolves, or, alternatively, how different user communities reinterpret the same concept in a different context so that the risk is having these concepts lose their rhetorical, descriptive and applicative power~\cite{cabitza2015boundary}. In a more general sense, the topic is important beyond its linguistic implications, especially for managing semantic interoperability for federations; respective research to date has focused on the generation of semantic mappings and has tended to ignore the problem of dealing with the dynamism of both the data and the schemata that is characteristic of real-world integration problems~\cite{brennan2014managed}.

%%%%%%%%%%%%%%%%%%%%%%%%%%%%%%%%%%%%%%%%%%%%%%%%%%%%%%%%%%%%%%%%%%%%%%%%%%%%%%%
\section{Methodology}
\begin{figure*}[htb!]
  \centering
  \includegraphics[width=0.7\textwidth]{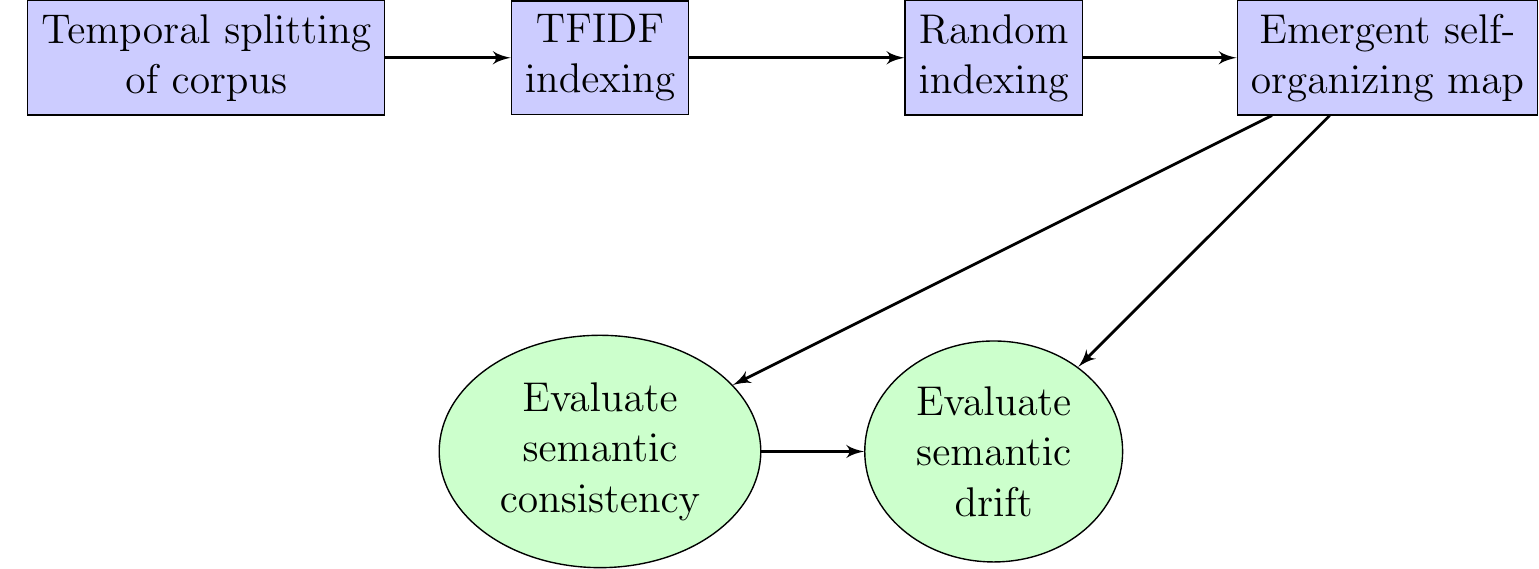}
  \caption{Outline of methodology.}
\label{outline}
\end{figure*}

We build a vector space model by random indexing that is able to closely track the changes of an evolving text collection. We project the space to a two-dimensional surface where clusters and shifts are more apparent by emergent self-organizing maps; this projection preserves the local topology of the high-dimensional space and allows us to model dynamic semantic fields. We use WordNet-based referential similarity measures to evaluate semantic consistency and also to detect semantic drifts over time. Refer to Figure~\ref{outline} for an outline.

%%%%%%%%%%%%%%%%%%%%%%%%%%%%%%%%%%%%%%%%%%%%%%%%%%%%%%%%%%%%%%%%%%%%%%%%%%%%%%
\subsection{Distributional similarity and random indexing}
We build a TFIDF vector space model of the corpus, which provides the foundation for most distributional semantic distance measures. The basic TFIDF space is known to be extremely sparse, having 1--5~\% nonzero elements. Latent semantic analysis, or latent semantic indexing~\cite{deerwester90:lsi} measures semantic information through co-occurrence analysis in the corpus, but it reduces the dimensionality and solves the problem of sparsity. The dimension of the vector space is reduced by singular value decomposition.

Random indexing is a similar idea which does not rely on the use of computationally intensive matrix decomposition. This makes random indexing a much more scalable technique in practice. Instead of first constructing a huge co-occurrence matrix and then using a separate dimension reduction phase, random projection builds an incremental word space model~\cite{kanerva2000random}. The random projection technique can be described as a three-step operation:
\begin{itemize}
\item First, each document in the corpus is assigned a unique and randomly generated representation called an index vector. These index vectors are sparse, high-dimensional, and ternary, which means that their dimensionality ($d$) is on the order of hundreds, and that they consist of a small number of randomly distributed values, with the rest of the elements of the vectors set to 0.
\item Then, context vectors are produced by scanning through the text,  and each time a word occurs in a context (e.g. in a document, or within a sliding context window), that context's $d$-dimensional index vector is added to the context vector for the word in question. Words are thus represented by $d$-dimensional context vectors that are effectively the sum of the words' contexts.
\end{itemize}
Comparing the term vectors in the random indexed space by a similarity measure such as the Euclidean distance or cosine dissimilarity enables a quantitative framework for semantic analysis.

%%%%%%%%%%%%%%%%%%%%%%%%%%%%%%%%%%%%%%%%%%%%%%%%%%%%%%%%%%%%%%%%%%%%%%%%%%%%%%%
\subsection{Semantic fields in emergent self-organizing maps}
The vectors of the term space are disjoint locations in a high-dimensional space. A field, on the other hand, is defined at all points in a space. To bridge this problem, we embed the vector space on a two-dimensional surface using emergent self-organizing maps~\cite{wittek2014vector}. 

A self-organizing map is a two-dimensional grid of artificial neurons. Each neuron is associated with a weight vector that matches the dimension of the training data. We take an instance of the training data, find the closest weight vector, and pull it closer to the data instance. We also pull the weight vectors of nearby neurons closer to the data instance, with decreasing weight as we get further from the best matching unit. We repeat this procedure with every training instance. This constitutes one epoch. We repeat the same process in the second epoch, but with a smaller neighbourhood radius, and a lower learning rate when adjusting the weight vectors. Eventually, the neighbourhood function decreases to an extent that training might stop. The time needed to train an SOM grows linearly with the data set size, and it grows linearly with the number of neurons in the SOM\@. The resulting network reflects the local topology of the high-dimensional space~\cite{kohonen2001som}.  Emergent self-organizing maps contain a much larger number of target nodes for embedding, and thus capture the topology of the original space more accurately~\cite{ultsch2005esom}. Using a toroid map avoids edge effects. 

Some nodes of the emergent self-organizing map correspond to one or more terms; these nodes or best matching units have a special role as they identify semantic content with one or more terms. The rest of the nodes act as an interpolation of the semantic field. Since the field is continuous in nature, we use toroid maps -- a planar map would introduce an artificial discrete cut-off at the edges.

%%%%%%%%%%%%%%%%%%%%%%%%%%%%%%%%%%%%%%%%%%%%%%%%%%%%%%%%%%%%%%%%%%%%%%%%%%%%%%%
\subsection{WordNet-based similarity metrics and validating consistency}
\label{WN_metrics}
WordNet (WN) is a large lexical database of English, created and maintained by Princeton University~\cite{miller1995wordnet}. It is publicly available to research and commercial users free of charge and its latest version is 3.1 (released Nov'12). WordNet's popularity arguably lies on the fact that, besides merely offering short definitions and usage examples of the contained nouns, verbs, adverbs and adjectives, it also introduces certain types of semantic relations between terms. Examples of such relations include synonymy, hyper/hyponymy (\emph{is}-a relationship), meronymy (part-of relationship) etc. WN can, thus, be viewed as a combination of a dictionary and a thesaurus.

Due to its structure described above, WN has been extensively deployed in tasks related to determining the semantic similarity between terms, primarily in automatic text analysis and artificial intelligence applications. Towards this direction, many semantic similarity metrics have been proposed, which can be grouped into four different categories~\cite{meng2013review}: path-based, content-based, feature-based and hybrid metrics.

\begin{itemize}
  \item In \emph{path-based metrics} the similarity between two terms depends on their relative position in the WN taxonomy as well as on the length of the path linking the concepts. Representative examples deploying path-based measures include~\cite{bulskov2002measuring},~\cite{leacock1998combining},~\cite{li03semsimi} and~\cite{wu1994semantics}.
  
  \item \emph{Content-based metrics} are based on the information content available for each concept in WN\@. The more common information two concepts share, the more similar they are. Examples belonging to this category include~\cite{jiang1997ssb},~\cite{lin1998inf},~\cite{Lord03} and~\cite{resnik1995uic}.
  
  \item \emph{Feature-based metrics} are based on the properties of the WN ontology for obtaining a similarity value. The more common (and the fewer non-common) characteristics two concepts have (e.g. their definitions or ``glosses'' in WN), the more similar they are. Relationships to other similar terms in the taxonomy are also taken into consideration. Some related approaches are the classical model proposed by Tversky~\cite{tversky77featuressimilarity} or the more recent approach presented in~\cite{sanchez2012ontology}.
  
  \item Finally, the \emph{hybrid metrics} combine the ideas presented above. The following constitute paradigms of applying hybrid similarity metrics:~\cite{dong2009hybrid},~\cite{Rodriguez03determiningsemantic} and~\cite{zhou2008new}.
\end{itemize}

Apart from the above distinction, Varelas et al. also distinguish semantic similarity measures in \emph{single ontology} and \emph{cross ontology} methods~\cite{varelas2005semantic}; the former assume that the terms compared all come from the same reference ontology, while the latter compare terms from two different ontologies. Since it is not easy to directly compare the structure and information content of different ontologies, the case of cross ontology similarity typically employs hybrid or feature-based methods (e.g. see~\cite{Rodriguez03determiningsemantic} and~\cite{li03semsimi}).

%%%%%%%%%%%%%%%%%%%%%%%%%%%%%%%%%%%%%%%%%%%%%%%%%%%%%%%%%%%%%%%%%%%%%%%%%%%%%%
\subsection{Quantifying drifts}
The foundation for measuring the drift is based on random indexing of subsequent TFIDF spaces with a fixed random seed. By following this method, we are able to derive subsequent low-dimensional spaces which can be compared against one another. We train an emergent self-organizing map on each of these term vector spaces: after the first period, we continue training the map with a lower learning rate to arrive at smoothly changing dynamics.

As a potential source of confusion, we point out that the time-like variable of the iterations and the epochs in the training of the ESOM are unrelated to the temporal aspects of the corpus. By time we always refer to the time related to the corpus, and by a period we refer to documents belonging to a certain time interval of the corpus. Epoch, on the other hand, refers to the training rounds of the ESOM.

%%%%%%%%%%%%%%%%%%%%%%%%%%%%%%%%%%%%%%%%%%%%%%%%%%%%%%%%%%%%%%%%%%%%%%%%%%%%%%
\section{Experiment design}
The experiments were based on a large text corpus on which the TFIDF spaces and random indices were created. The random indices were used to generate the sequence of self-organizing maps. We analysed the topology of these maps for consistency and drifts.

%%%%%%%%%%%%%%%%%%%%%%%%%%%%%%%%%%%%%%%%%%%%%%%%%%%%%%%%%%%%%%%%%%%%%%%%%%%%%%
\subsection{Corpus}
Book reviews are the literary genre equivalent of abstracts to scientific articles, cross-pollinated by the idea of crowd-sourcing underlying recommender systems~\cite{resnick1997recommender}, i.e.\ one summary per article produced per one professional abstracting and indexing service vs. potentially many summaries of the same item, written by users as part professional, part lay contributors. In a sense both approaches represent user feedback. From a methodological perspective, due to the nature of condensed semantic content in them, book reviews processing falls in the category of e.g. text summarization~\cite{chatterjee2015random} on the one hand, combined with sentiment analysis~\cite{tang2015sentiment} on the other hand. Due to this blend, they represent an interesting and scalable resource of complex semantic content for ``neuromorphic'' studies.

The experiments described here were based on Stanford's Amazon book reviews data set~\cite{mcauley2013hidden}, which is publicly available as part of the University's SNAP project\footnote{\url{http://snap.stanford.edu/index.html}, last accessed: Jan'15.}. The data set spanned a period of 18 years and included approximately 12.8M book reviews up to March 2013. Every item in the data set included product and user information, ratings, as well as a plain text content description. The collection contained 51 degenerate time stamps, the corresponding instances were discarded.

\begin{table}[htb!]
  \begin{center}
     \begin{tabular}{ccc}
        \textbf{Period 1} & \textbf{Period 2} & \textbf{Period 3}\\
         Until 30 Jan 2003 & Until 03 Aug 2008 & Until 04 Mar 2013 \\
         45162 terms & 49400 terms & 50672 terms
     \end{tabular}
  \end{center}
\caption{Key statistics of the temporal split of the corpus.}
\label{periodstats}
\end{table}

We split the corpus in three periods, each containing close to 4.3M reviews. The key characteristics are summarised in Table~\ref{periodstats}.

\begin{table*}[htb!]
\begin{center}
    \begin{tabular}{|c|ccc|ccc|ccc|}
    \hline
    % after \\: \hline or \cline{col1-col2} \cline{col3-col4} \ldots
    &\multicolumn{3}{c|}{Period 1}
    &\multicolumn{3}{c|}{Period 2}   
    &\multicolumn{3}{c|}{Period 3} \\ \hline
    Term count &  $a=0.05$ & $a=0.1$ & $a=0.2$ & $a=0.05$ & $a=0.1$ & $a=0.2$ & $a=0.05$ & $a=0.1$ & $a=0.2$ \\ \hline
    $N\geq3$   & 0.287 & 0.334 & 0.404 & 0.269 & 0.320 & 0.406 & 0.259 & 0.316 & 0.400\\
    $N\geq5$   & 0.339 & 0.380 & 0.429 & 0.327 & 0.366 & 0.421 & 0.304 & 0.353 & 0.410\\
    $N\geq10$  & 0.383 & 0.416 & 0.453 & 0.372 & 0.401 & 0.448 & 0.332 & 0.380 & 0.431\\ \hline
    \end{tabular}
\end{center}
        \caption{Percentages of rejection of the null hypothesis for the \emph{t}-test for each period and for different neuron populations.}
\label{table1}
\end{table*}

\begin{table*}[htb!]
\begin{center}
    \begin{tabular}{|c|ccc|}
    \hline
    % after \\: \hline or \cline{col1-col2} \cline{col3-col4} \ldots
    Periods &  $a=0.05$ & $a=0.1$ & $a=0.2$ \\ \hline
    $3-p1-p2$   & 0.051 & 0.130 & 0.860 \\
    $3-p2-p3$   & 0.211 & 0.614 & 0.505 \\
    $5-p1-p2$  & 0.330 & 0.251 & 0.515 \\ 
    $5-p2-p3$  & 0.051 & 0.301 & 0.351 \\ 
    $10-p1-p2$  & 0.318 & 0.494 & 0.833 \\ 
    $10-p2-p3$  & 0.060 & 0.344 & 0.451 \\ \hline
    \end{tabular}
\end{center}
        \caption{p-values of the comparison of percentages displayed in Table~\ref{table1} between consecutive periods and for different neuron populations.}
\label{table2}
\end{table*}

%%%%%%%%%%%%%%%%%%%%%%%%%%%%%%%%%%%%%%%%%%%%%%%%%%%%%%%%%%%%%%%%%%%%%%%%%%%%%%%
\subsection{Computational background}
We used Lucene\footnote{\url{http://lucene.apache.org/}, last accessed: Jan'15}, which is an information retrieval software library that builds an inverted index, which can be interpreted as a row-major sparse representation of a term-document matrix. We used the SemanticVectors package~\cite{widdows2008semantic} for reducing the dimensionality of the space. For training the emergent self-organizing maps, we used Somoclu~\cite{wittek2013somoclu}.

The implementation of the semantic similarity metrics was based on WS4J\footnote{\url{https://code.google.com/p/ws4j/}, last accessed: Jan'15.}, a Java API for several of the semantic relatedness/similarity algorithms presented in section~\ref{WN_metrics}. WS4J works over WordNet 3.0 and constitutes an improved version of the Perl-based WordNet-Similarity-2.05\footnote{\url{http://wn-similarity.sourceforge.net/}, last accessed: Feb'15.}. In the experiments described subsequently, we deployed a representative path-based semantic similarity method (Wu and Palmer's~\cite{wu1994semantics}), while in the imminent future we plan to investigate the behaviour of more methods (path-based and content-based). Note that all experiments are open source\footnote{\url{https://github.com/peterwittek/concept_drifts}}.

%%%%%%%%%%%%%%%%%%%%%%%%%%%%%%%%%%%%%%%%%%%%%%%%%%%%%%%%%%%%%%%%%%%%%%%%%%%%%
\subsection{Consistency in a single time period}
This subsection is aimed at assessing whether proximity of terms in the toroid plane indicates a strong underlying semantic similarity. Our focus lied on neurons that had more than 1 term assigned to them and we were interested in the average similarities within these neurons. Towards this direction and in order to evaluate the consistency of our approach in the case of a single time period, initially we randomly divided the terms of the period into clusters of 5 terms each (5 is the average count of terms assigned to non-empty neurons in the toroid plane). For each of the groups, we computed the average semantic similarity; then, based on all the derived average similarities, we determined their empirical probability distribution. The latter was a good approximation of a normal distribution. This enabled us to make use of the \emph{t}-test for evaluating the significance of the similarity, as described subsequently. 

The key idea was that we assumed that the terms within each neuron constituted a random group from the total population. Hence, the average similarities within the neurons followed the normal distribution with mean equal to the empirical mean ($\mu_0$) from the above distribution and variance equal to its empirical variance ($\sigma_{0}^{2}$). Based on this assumption ($H_0: \mu=\mu_0$), we performed a 1-sided \emph{t}-test for a predefined level of significance ($a$), in order to assess whether the average within each neuron was statistically significantly greater than this empirical mean ($H_1: \mu>\mu_0$). We considered three generic cases: (a) neurons containing 3 or more terms, (b) neurons containing 5 or more terms, and, finally, (c) neurons containing 10 or more terms. For each of the three cases, we repeatedly performed the 1-sided \emph{t}-test for every neuron and calculated the percentage ($\hat{p}$) of the cases where we could reject the null hypothesis. A percentage greater than $a$ indicated that the number of samples with high average similarity was much greater than expected, based on the assumption that the average similarities followed the normal distribution $N(\mu_0,\sigma_{0}^{2})$.

Consequently, in order to assess whether this percentage was indeed anticipated or not, we performed a one-sided binomial test ($H_0: \hat{p}=a, H_1: \hat{p}>a$) at $0.05$ level of significance, for comparing this percentage with the level of significance of the \emph{t}-test. In the cases when $H_0$ was rejected, we deduced that both $\hat{p}$ and the overall level of similarity of terms \emph{within the neurons} were statistically significantly greater than expected, based on our key initial assumption. This meant that terms within a neuron demonstrated greater similarity in comparison to a random group of terms, indicating that this grouping made sense. Table~\ref{table1} displays the percentages $\hat{p}$ for each of the three cases and for different levels of significance for each of the three periods (see next subsection). The p-values derived from the application of the binomial test are not displayed, since they are all negligible. This holds because the percentage is much greater than $a$ in every case.

As an example, consider the following neurons: $ex_1 = \{\mathrm{thymine}, 
\mathrm{cytosine}, \mathrm{uracil}\}$ and $ex_2 = \{\mathrm{pair}, \mathrm{harvard}, \mathrm{scorpio}, \mathrm{monsignor}, \mathrm{misrepresentation}\}$. Intuitively, $ex_1$ displays apparent semantic similarity (all terms are nucleobases), in contrast to the terms included in $ex_2$ that do not. We would expect that $H_0: \mu=\mu_0$ would be rejected for $ex_1$ and not rejected for $ex_2$, which is indeed the case. However, there are some cases when the semantic similarity of terms within a neuron is not that apparent, but the null hypothesis is again rejected. This is due to the chosen path-based semantic similarity metric~\cite{wu1994semantics}; thus, we leave as future work the application of further path-based and information content-based methods. All in all, the robustness of this statistical approach is based on our initial assumption that the population of averages followed a normal distribution which we tested empirically. However, we believe this is the groundwork for interesting future investigations.

%%%%%%%%%%%%%%%%%%%%%%%%%%%%%%%%%%%%%%%%%%%%%%%%%%%%%%%%%%%%%%%%%%%%%%%%%%%%%
\subsection{Dynamics of semantic consistency}
We were interested to find out if over several periods the overall similarity within each neuron converges. Our experiment in this paper involved three periods and for each of them we repeated the process described in the previous subsection. We calculated the percentages of the cases where we rejected the null hypothesis of the \emph{t}-test for all three periods, as displayed in Table~\ref{table1}. As observed in the table, there is a slight decrease of the percentages from period to period. To assess whether this decrease is important, we performed a test for proportional comparison at $a=0.05$. The resulting p-values are shown in Table~\ref{table2} and indicate that this decrease is not statistically significant in every case. This means that there are indeed some slight differences that do not demonstrate divergence. More periods would be needed in order to investigate whether there is convergence, which we will study in the imminent future.

%%%%%%%%%%%%%%%%%%%%%%%%%%%%%%%%%%%%%%%%%%%%%%%%%%%%%%%%%%%%%%%%%%%%%%%%%%%%%
\section{Conclusion and future work}
Recently increased attention has been paid to models of evolving
semantic content, something we represented as a dynamic vector field.
This approach was based on the hypothesis of semantic continuity in the
vocabulary, allowing both for manifest (actual) and latent (potential)
word content mapped to lexical forms, with word meaning in term space
behaving like ``energy'' while constructing conceptual categories. In
spite of the simple design to analyse book reviews over 18 years in just
three periods, the \emph{t}-test confirmed that the overall level of similarity
of terms within the ESOM grid neurons was significantly higher than
expected, i.e.\ neuron content was statistically consistent.

We plan to continue this line of research in different directions,
including the following:
\begin{itemize}
\item Increase the number of periods to compute more detailed visual maps
over shorter time spans;
\item Increase grid granularity to reduce term overlap on neurons;
\item Upgrade this model by introducing reflexive random indexing to
smoothen transition between periods;
\item Interpret the vector field model in terms of e.g. process
philosophy.
\end{itemize}

%%%%%%%%%%%%%%%%%%%%%%%%%%%%%%%%%%%%%%%%%%%%%%%%%%%%%%%%%%%%%%%%%%%%%%%%%%%%%%%
\section{Acknowledgement}
This project has received funding from the European Union's Seventh Framework Programme for research, technological development and demonstration under grant agreement no FP7-601138 PERICLES\@. S. Dar\'anyi and P. Wittek are grateful to Dominic Widdows (Microsoft Bing) and Trevor Cohen (University of Houston) for assistance with the Semantic Vectors package. The authors would also like to thank Julian McAuley (Stanford University) for granting access to the Amazon data set.

%%%%%%%%%%%%%%%%%%%%%%%%%%%%%%%%%%%%%%%%%%%%%%%%%%%%%%%%%%%%%%%%%%%%%%%
\bibliographystyle{IEEEtranS}
\bibliography{bibliography}

% Generated by IEEEtranS.bst, version: 1.12 (2007/01/11)
\begin{thebibliography}{10}
\providecommand{\url}[1]{#1}
\csname url@samestyle\endcsname
\providecommand{\newblock}{\relax}
\providecommand{\bibinfo}[2]{#2}
\providecommand{\BIBentrySTDinterwordspacing}{\spaceskip=0pt\relax}
\providecommand{\BIBentryALTinterwordstretchfactor}{4}
\providecommand{\BIBentryALTinterwordspacing}{\spaceskip=\fontdimen2\font plus
\BIBentryALTinterwordstretchfactor\fontdimen3\font minus
  \fontdimen4\font\relax}
\providecommand{\BIBforeignlanguage}[2]{{%
\expandafter\ifx\csname l@#1\endcsname\relax
\typeout{** WARNING: IEEEtranS.bst: No hyphenation pattern has been}%
\typeout{** loaded for the language `#1'. Using the pattern for}%
\typeout{** the default language instead.}%
\else
\language=\csname l@#1\endcsname
\fi
#2}}
\providecommand{\BIBdecl}{\relax}
\BIBdecl

\bibitem{amit1992modeling}
D.~J. Amit, \emph{Modeling Brain Function: The World of Attractor Neural
  Networks}.\hskip 1em plus 0.5em minus 0.4em\relax Cambridge University Press,
  1992.

\bibitem{antoniou2011semantic}
G.~Antoniou, M.~d'Aquin, and J.~Z. Pan, ``Semantic web dynamics,'' \emph{Web
  Semantics: Science, Services and Agents on the World Wide Web}, vol.~9,
  no.~3, pp. 245--246, 2011.

\bibitem{baker08languageChange}
A.~Baker, ``Computational approaches to the study of language change,''
  \emph{Language and Linguistics Compass}, vol.~2, no.~3, pp. 289--307, 2008.

\bibitem{baroni2010distributional}
M.~Baroni and A.~Lenci, ``Distributional memory: A general framework for
  corpus-based semantics,'' \emph{Computational Linguistics}, vol.~36, no.~4,
  pp. 673--721, 2010.

\bibitem{beeferman1997mod}
D.~Beeferman, A.~Berger, and J.~Lafferty, ``A model of lexical attraction and
  repulsion,'' in \emph{Proceedings of ACL-97, 35th Annual Meeting of the
  Association for Computational Linguistics}, July 1997, pp. 373--380.

\bibitem{blacoe2013quantum}
W.~Blacoe, E.~Kashefi, and M.~Lapata, ``A quantum-theoretic approach to
  distributional semantics,'' in \emph{Proceedings of NAACL-HLT-13, Conference
  of the North American Chapter of the Association for Computational
  Linguistics: Human Language Technologies}, June 2013, pp. 847--857.

\bibitem{brennan2014managed}
R.~Brennan, B.~Walshe, and D.~O'Sullivan, ``Managed semantic interoperability
  for federations,'' \emph{Journal of Network and Systems Management}, vol.~22,
  no.~3, pp. 302--330, 2014.

\bibitem{bruce2001biomedical}
E.~Bruce, \emph{Biomedical signal processing and signal modeling}.\hskip 1em
  plus 0.5em minus 0.4em\relax Wiley-Interscience, 2001.

\bibitem{bulskov2002measuring}
H.~Bulskov, R.~Knappe, and T.~Andreasen, ``On measuring similarity for
  conceptual querying,'' in \emph{Flexible Query Answering Systems}, ser.
  Lecture Notes in Computer Science, T.~Andreasen, A.~Motro, H.~Christiansen,
  and H.~L. Larsen, Eds.\hskip 1em plus 0.5em minus 0.4em\relax Springer, 2002,
  vol. 2522, pp. 100--111.

\bibitem{bylesjo2006opls}
M.~Bylesj{\"o}, M.~Rantalainen, O.~Cloarec, J.~K. Nicholson, E.~Holmes, and
  J.~Trygg, ``{OPLS} discriminant analysis: Combining the strengths of {PLS-DA}
  and {SIMCA} classification,'' \emph{Journal of Chemometrics}, vol.~20, no.
  8-10, pp. 341--351, 2006.

\bibitem{cabitza2015boundary}
F.~Cabitza, ``At the boundary of communities and roles: Boundary objects and
  knowledge artifacts as resources for {IS} design,'' in \emph{From Information
  to Smart Society}.\hskip 1em plus 0.5em minus 0.4em\relax Springer, 2015, pp.
  149--160.

\bibitem{chatterjee2015random}
N.~Chatterjee and P.~K. Sahoo, ``Random indexing and modified random indexing
  based approach for extractive text summarization,'' \emph{Computer Speech \&
  Language}, vol.~29, no.~1, pp. 32--44, 2015.

\bibitem{cohen2012discovery}
T.~Cohen, D.~Widdows, R.~W. Schvaneveldt, and T.~C. Rindflesch, ``Discovery at
  a distance: Farther journeys in predication space,'' in \emph{Proceedings of
  BIBMW-12, IEEE International Conference on Bioinformatics and Biomedicine
  Workshops}, 2012, pp. 218--225.

\bibitem{cooper1999linguistic}
D.~L. Cooper, \emph{Linguistic attractors: The cognitive dynamics of language
  acquisition and change}.\hskip 1em plus 0.5em minus 0.4em\relax John
  Benjamins Publishing, 1999, vol.~2.

\bibitem{daranyi12connecting}
S.~Dar\'anyi and P.~Wittek, ``Connecting the dots: Mass, energy, word meaning,
  and particle-wave duality,'' in \emph{Proceedings of QI-12, 6th International
  Quantum Interaction Symposium}, June 2012.

\bibitem{desaussure1916course}
F.~{De Saussure}, \emph{Course in General Linguistics}, 1916.

\bibitem{deerwester90:lsi}
S.~Deerwester, S.~Dumais, G.~Furnas, T.~Landauer, and R.~Harshman, ``Indexing
  by latent semantic analysis,'' \emph{Journal of the American Society for
  Information Science}, vol.~41, no.~6, pp. 391--407, 1990.

\bibitem{delany2005case}
S.~J. Delany, P.~Cunningham, A.~Tsymbal, and L.~Coyle, ``A case-based technique
  for tracking concept drift in spam filtering,'' \emph{Knowledge-Based
  Systems}, vol.~18, no.~4, pp. 187--195, 2005.

\bibitem{deo2015diachronic}
A.~Deo, ``Diachronic semantics,'' \emph{Annual Review of Linguistics}, vol.~1,
  no.~1, pp. 179--197, 2015.

\bibitem{dong2009hybrid}
H.~Dong, F.~K. Hussain, and E.~Chang, ``A hybrid concept similarity measure
  model for ontology environment,'' in \emph{On the Move to Meaningful Internet
  Systems: OTM 2009 Workshops}, ser. Lecture Notes in Computer Science,
  R.~Meersman, P.~Herrero, and T.~S. Dillon, Eds., vol. 5872.\hskip 1em plus
  0.5em minus 0.4em\relax Springer, 2009, pp. 848--857.

\bibitem{elman2004alternative}
J.~L. Elman, ``An alternative view of the mental lexicon,'' \emph{Trends in
  Cognitive Sciences}, vol.~8, no.~7, pp. 301--306, 2004.

\bibitem{erk2013measuring}
K.~Erk, D.~McCarthy, and N.~Gaylord, ``Measuring word meaning in context,''
  \emph{Computational Linguistics}, vol.~39, no.~3, pp. 511--554, 2013.

\bibitem{erk2008structured}
K.~Erk and S.~Pad{\'o}, ``A structured vector space model for word meaning in
  context,'' in \emph{Proceedings of EMNLP-08, 13th Conference on Empirical
  Methods in Natural Language Processing}, October 2008, pp. 897--906.

\bibitem{fodor1975language}
J.~A. Fodor, \emph{The Language of Thought}.\hskip 1em plus 0.5em minus
  0.4em\relax Harvard University Press, 1975, vol.~5.

\bibitem{frege1892sb}
G.~Frege, ``Sense and reference,'' \emph{The Philosophical Review}, vol.~57,
  no.~3, pp. 209--230, 1948.

\bibitem{gallant91}
S.~I. Gallant, ``A practical approach for representing context and for
  performing word sense disambiguation using neural networks,'' \emph{Neural
  Computation}, vol.~3, pp. 293--309, 1991.

\bibitem{gliozzo2009semantic}
A.~Gliozzo and C.~Strapparava, \emph{Semantic domains in computational
  linguistics}.\hskip 1em plus 0.5em minus 0.4em\relax Springer Science \&
  Business Media, 2009.

\bibitem{gliozzo2007domain}
A.~M. Gliozzo, M.~Pennacchiotti, and P.~Pantel, ``The domain restriction
  hypothesis: Relating term similarity and semantic consistency,'' in
  \emph{Proceedings of NAACL-HLT-07, Conference of the North American Chapter
  of the Association for Computational Linguistics: Human Language
  Technologies}, April 2007, pp. 131--138.

\bibitem{goncalves2013rcd}
P.~M. Gon{\c{c}}alves~Jr and R.~S. M.~d. Barros, ``Rcd: A recurring concept
  drift framework,'' \emph{Pattern Recognition Letters}, vol.~34, no.~9, pp.
  1018--1025, 2013.

\bibitem{grefenstette2013multi}
E.~Grefenstette, G.~Dinu, Y.-Z. Zhang, M.~Sadrzadeh, and M.~Baroni,
  ``Multi-step regression learning for compositional distributional
  semantics,'' \emph{arXiv:1301.6939}, 2013.

\bibitem{gulla2011concept}
J.~A. Gulla, G.~Solskinnsbakk, P.~Myrseth, V.~Haderlein, and O.~Cerrato,
  ``Concept signatures and semantic drift,'' in \emph{Web Information Systems
  and Technologies}.\hskip 1em plus 0.5em minus 0.4em\relax Springer, 2011, pp.
  101--113.

\bibitem{hansemantic}
X.~Han and L.~Sun, ``Semantic consistency: A local subspace based method for
  distant supervised relation extraction,'' in \emph{Proceedings of ACL-14,
  52nd Annual Meeting of the Association for Computational Linguistics}, June
  2014, pp. 718--724.

\bibitem{harris70:distr}
Z.~Harris, ``Distributional structure,'' in \emph{Papers in structural and
  transformational Linguistics}, ser. Formal Linguistics, Z.~Harris, Ed.\hskip
  1em plus 0.5em minus 0.4em\relax Humanities Press, 1970, pp. 775--794.

\bibitem{house2000linguistic}
J.~House, ``Linguistic relativity and translation,'' \emph{Amsterdam Studies in
  the Theory and History of Linguistic Science}, vol.~4, pp. 69--88, 2000.

\bibitem{jiang1997ssb}
J.~Jiang and D.~Conrath, ``Semantic similarity based on corpus statistics and
  lexical taxonomy,'' in \emph{Proceedings of ROCLING-97, International
  Conference on Research in Computational Linguistics}, 1997, pp. 19--33.

\bibitem{kanerva2000random}
P.~Kanerva, J.~Kristofersson, and A.~Holst, ``Random indexing of text samples
  for latent semantic analysis,'' in \emph{Proceedings of CogSci-00, 22nd
  Annual Conference of the Cognitive Science Society}, vol. 1036, 2000.

\bibitem{karlgren2001wu}
J.~Karlgren and M.~Sahlgren, ``From words to understanding,'' in
  \emph{Foundations of Real-World Intelligence}, Y.~Uesaka, P.~Kanerva, and
  H.~Asoh, Eds.\hskip 1em plus 0.5em minus 0.4em\relax CSLI Publications, 2001,
  pp. 294--308.

\bibitem{kohonen2001som}
T.~Kohonen, \emph{Self-Organizing Maps}.\hskip 1em plus 0.5em minus 0.4em\relax
  Springer, 2001.

\bibitem{kontostathis2006ful}
A.~Kontostathis and W.~Pottenger, ``A framework for understanding latent
  semantic indexing ({LSI}) performance,'' \emph{Information Processing and
  Management}, vol.~42, no.~1, pp. 56--73, 2006.

\bibitem{lauriston1995criteria}
A.~Lauriston, ``Criteria for measuring term recognition,'' in \emph{Proceedings
  of EACL-95, 7th Conference of the European Chapter of the Association for
  Computational Linguistics}, March 1995, pp. 17--22.

\bibitem{leacock1998combining}
C.~Leacock and M.~Chodorow, ``Combining local context and {WordNet} similarity
  for word sense identification,'' in \emph{{Wordnet}: An Electronic Lexical
  Database}, C.~Fellfaum, Ed.\hskip 1em plus 0.5em minus 0.4em\relax MIT Press,
  1998, pp. 265--283.

\bibitem{lesk1986}
M.~Lesk, ``Automatic sense disambiguation using machine readable dictionaries:
  How to tell a pine cone from an ice cream cone?'' in \emph{Proceedings of
  SIGDOC-86, 5th Annual International Conference on Systems Documentation},
  1986, pp. 24--26.

\bibitem{levine2009quantum}
I.~N. Levine and P.~Learning, \emph{Quantum Chemistry}.\hskip 1em plus 0.5em
  minus 0.4em\relax Pearson Prentice Hall, 2009, vol.~6.

\bibitem{li03semsimi}
Y.~Li, Z.~Bandar, and S.~McLean, ``An approach for measuring semantic
  similarity between words using multiple information sources,'' \emph{IEEE
  Transactions on Knowledge and Data Engineering}, vol.~15, no.~4, pp.
  871--882, July 2003.

\bibitem{liao2009isorankn}
C.-S. Liao, K.~Lu, M.~Baym, R.~Singh, and B.~Berger, ``{IsoRankN}: Spectral
  methods for global alignment of multiple protein networks,''
  \emph{Bioinformatics}, vol.~25, no.~12, pp. i253--i258, 2009.

\bibitem{lin1998inf}
D.~Lin, ``An information-theoretic definition of similarity,'' in
  \emph{Proceedings of ICML-98, 15th International Conference on Machine
  Learning}, June 1998, pp. 296--304.

\bibitem{Lord03}
P.~W. Lord, R.~D. Stevens, A.~Brass, and C.~A. Goble, ``Investigating semantic
  similarity measures across the {Gene Ontology}: the relationship between
  sequence and annotation.'' \emph{Bioinformatics}, vol.~19, no.~10, pp.
  1275--1283, 2003.

\bibitem{mcauley2013hidden}
J.~McAuley and J.~Leskovec, ``Hidden factors and hidden topics: Understanding
  rating dimensions with review text,'' in \emph{Proceedings of RecSys-13, 7th
  ACM Conference on Recommender Systems}.\hskip 1em plus 0.5em minus
  0.4em\relax ACM, 2013, pp. 165--172.

\bibitem{meng2013review}
L.~Meng, R.~Huang, and J.~Gu, ``A review of semantic similarity measures in
  {WordNet},'' \emph{International Journal of Hybrid Information Technology},
  vol.~6, no.~1, pp. 1--12, January 2013.

\bibitem{mezey1987potential}
P.~G. Mezey, \emph{Potential Energy Hypersurfaces}.\hskip 1em plus 0.5em minus
  0.4em\relax Elsevier Amsterdam, 1987.

\bibitem{miller1995wordnet}
G.~A. Miller, ``{WordNet}: A lexical database for english,''
  \emph{Communinactions of the ACM}, vol.~38, no.~11, pp. 39--41, November
  1995.

\bibitem{mohammad2005dmp}
S.~Mohammad and G.~Hirst, ``Distributional measures as proxies for semantic
  relatedness,'' 2005, submitted for publication.

\bibitem{morris2003tra}
J.~Morris, C.~Beghtol, and G.~Hirst, ``Term relationships and their
  contribution to text semantics and information literacy through lexical
  cohesion,'' in \emph{Proceedings of CAIS-03, 31st Annual Conference of the
  Canadian Association for Information Science}, May 2003.

\bibitem{morris1991lcc}
J.~Morris and G.~Hirst, ``Lexical cohesion computed by thesaural relations as
  an indicator of the structure of text,'' \emph{Computational Linguistics},
  vol.~17, no.~1, pp. 21--48, 1991.

\bibitem{pado2007dependency}
S.~Pad{\'o} and M.~Lapata, ``Dependency-based construction of semantic space
  models,'' \emph{Computational Linguistics}, vol.~33, no.~2, pp. 161--199,
  2007.

\bibitem{peat1991ltc}
H.~Peat and P.~Willett, ``The limitations of term co-occurrence data for query
  expansion in document retrieval systems,'' \emph{Journal of the American
  Society for Information Science}, vol.~42, no.~5, pp. 378--383, 1991.

\bibitem{renouf2007search}
A.~Renouf and J.~Banerjee, ``The search for repulsion: A new corpus analytical
  approach,'' \emph{Towards Multimedia in Corpus Studies}, vol.~2, 2007.

\bibitem{resnick1997recommender}
P.~Resnick and H.~R. Varian, ``Recommender systems,'' \emph{Communications of
  the ACM}, vol.~40, no.~3, pp. 56--58, 1997.

\bibitem{resnik1995uic}
P.~Resnik, ``Using information content to evaluate semantic similarity in a
  taxonomy,'' in \emph{Proceedings of IJCAI-95, 14th International Joint
  Conference on Artificial Intelligence}, vol.~1, August 1995, pp. 448--453.

\bibitem{Rodriguez03determiningsemantic}
M.~A. Rodr\'iguez and M.~J. Egenhofer, ``Determining semantic similarity among
  entity classes from different ontologies,'' \emph{IEEE Transactions on
  Knowledge and Data Engineering}, vol.~15, no.~2, pp. 442--456, 2003.

\bibitem{ross2012exponentially}
G.~J. Ross, N.~M. Adams, D.~K. Tasoulis, and D.~J. Hand, ``Exponentially
  weighted moving average charts for detecting concept drift,'' \emph{Pattern
  Recognition Letters}, vol.~33, no.~2, pp. 191--198, 2012.

\bibitem{sanchez2012ontology}
D.~S\'{a}nchez, M.~Batet, D.~Isern, and A.~Valls, ``Ontology-based semantic
  similarity: A new feature-based approach,'' \emph{Expert Systems with
  Applications}, vol.~39, no.~9, pp. 7718--7728, 2012.

\bibitem{schutze97}
H.~Sch\"utze and T.~Pedersen, ``A co-occurrence-based thesaurus and two
  applications to information retrieval,'' \emph{Information Processing and
  Management}, vol.~3, no.~33, pp. 307--318, 1997.

\bibitem{socher2012semantic}
R.~Socher, B.~Huval, C.~D. Manning, and A.~Y. Ng, ``Semantic compositionality
  through recursive matrix-vector spaces,'' in \emph{Proceedings of
  EMNLP-CoNLL-12, Joint Conference on Empirical Methods in Natural Language
  Processing and Computational Natural Language Learning}.\hskip 1em plus 0.5em
  minus 0.4em\relax Association for Computational Linguistics, 2012, pp.
  1201--1211.

\bibitem{taira2007field}
R.~Taira, V.~Bashyam, and H.~Kangarloo, ``A field theoretical approach to
  medical natural language processing,'' \emph{IEEE Transactions on Information
  Technology in Biomedicine}, vol.~11, no.~4, p. 364, 2007.

\bibitem{tang2015sentiment}
D.~Tang, ``Sentiment-specific representation learning for document-level
  sentiment analysis,'' in \emph{Proceedings of WSDM-15, 8th ACM International
  Conference on Web Search and Data Mining}.\hskip 1em plus 0.5em minus
  0.4em\relax ACM, 2015, pp. 447--452.

\bibitem{trier1934sprachliche}
J.~Trier, ``Das sprachliche feld,'' \emph{Neue Jahrbucher fur Wissenschaft und
  Jugendbildung}, vol.~10, pp. 428--449, 1934.

\bibitem{turney2010frequency}
P.~D. Turney and P.~Pantel, ``From frequency to meaning: Vector space models of
  semantics,'' \emph{Journal of Artificial Intelligence Research}, vol.~37,
  no.~1, pp. 141--188, 2010.

\bibitem{tversky77featuressimilarity}
A.~Tversky, ``Features of similarity,'' \emph{Psychological Review}, vol.~84,
  no.~4, pp. 327--352, July 1977.

\bibitem{ultsch2005esom}
A.~Ultsch and F.~M\"orchen, ``{ESOM-Maps}: Tools for clustering, visualization,
  and classification with emergent {SOM},'' Data Bionics Research Group,
  University of Marburg, Tech. Rep., 2005.

\bibitem{varelas2005semantic}
G.~Varelas, E.~Voutsakis, P.~Raftopoulou, E.~G. Petrakis, and E.~E. Milios,
  ``Semantic similarity methods in {WordNet} and their application to
  information retrieval on the web,'' in \emph{Proceedings of WIDM-05, 7th
  Annual ACM International Workshop on Web Information and Data
  Management}.\hskip 1em plus 0.5em minus 0.4em\relax ACM, 2005, pp. 10--16.

\bibitem{wang2001indexing}
C.~Wang and X.~Wang, ``Indexing very high-dimensional sparse and quasi-sparse
  vectors for similarity searches,'' \emph{The VLDB Journal}, vol.~9, no.~4,
  pp. 344--361, 2001.

\bibitem{wang2011concept}
S.~Wang, S.~Schlobach, and M.~Klein, ``Concept drift and how to identify it,''
  \emph{Web Semantics: Science, Services and Agents on the World Wide Web},
  vol.~9, no.~3, pp. 247--265, 2011.

\bibitem{widdows2008semantic}
D.~Widdows and K.~Ferraro, ``Semantic vectors: a scalable open source package
  and online technology management application,'' in \emph{LREC-08, 6th
  International Conference on Language Resources and Evaluation}, May 2008.

\bibitem{widdows2014reasoning}
D.~Widdows and T.~Cohen, ``Reasoning with vectors: A continuous model for fast
  robust inference,'' \emph{Logic Journal of IGPL}, p. jzu028, 2014.

\bibitem{wittek2014vector}
P.~Wittek, S.~Dar\'anyi, and Y.~Lin, ``A vector field approach to lexical
  semantics,'' in \emph{To appear in Proceedings of QI-14, 8th International
  Conference on Quantum Interaction}, June 2014.

\bibitem{wittek2013somoclu}
P.~Wittek, S.~C. Gao, I.~S. Lim, and L.~Zhao, ``Somoclu: An efficient parallel
  library for self-organizing maps,'' \emph{arXiv:1305.1422}, 2015.

\bibitem{wu1994semantics}
Z.~Wu and M.~Palmer, ``Verb semantics and lexical selection,'' in
  \emph{Proceedings of ACL-94, 32nd Annual Meeting on Association for
  Computational Linguistics}.\hskip 1em plus 0.5em minus 0.4em\relax
  Association for Computational Linguistics, 1994, pp. 133--138.

\bibitem{yang2009citationrank}
L.~Yang, L.~Xu, and L.~He, ``A {CitationRank} algorithm inheriting google
  technology designed to highlight genes responsible for serious adverse drug
  reaction,'' \emph{Bioinformatics}, vol.~25, no.~17, pp. 2244--2250, 2009.

\bibitem{yuret1998discovery}
D.~Yuret, ``Discovery of linguistic relations using lexical attraction,''
  \emph{arXiv:cmp-lg/9805009}, 1998.

\bibitem{zhou2008new}
Z.~Zhou, Y.~Wang, and J.~Gu, ``New model of semantic similarity measuring in
  {Wordnet},'' in \emph{Proceedings of ISKE-08, 3rd International Conference on
  Intelligent System and Knowledge Engineering}, vol.~1, Nov 2008, pp.
  256--261.

\end{thebibliography}

\end{document}